\documentclass[sigconf]{acmart}
\usepackage{algorithm}
\usepackage{bbm}
\renewcommand\footnotetextcopyrightpermission[1]{} 
\AtBeginDocument{%
  }

\setcopyright{acmlicensed}
\copyrightyear{2018}
\acmYear{2018}
\acmDOI{XXXXXXX.XXXXXXX}

\acmISBN{978-1-4503-XXXX-X/18/06}

\usepackage{amsmath}
\usepackage{subfigure}
\usepackage{amsfonts}
\usepackage{bm}
\usepackage{multirow}
\usepackage{booktabs}  
\usepackage{makecell}
\usepackage{xcolor}
\usepackage{natbib}
\usepackage{makecell} 
\usepackage{enumitem}
\usepackage{subcaption}

\usepackage{graphicx}
\usepackage{color}
\usepackage{braket}
\usepackage{caption}
\usepackage{mathtools}
\usepackage{enumerate}
\usepackage{empheq}
\usepackage{calc}
\usepackage{dsfont}
\usepackage{balance}

\usepackage{colortbl}
\usepackage{algorithm}
\usepackage{algorithmic}

\begin{document}

\title{LLM-based Evaluation Policy Extraction for Ecological Modeling}





\author{Qi Cheng}
\affiliation{%
  \institution{University of Pittsburgh}
  \city{Pittsburgh}
  \state{PA}
  \country{USA}}
  \email{qic69@pitt.edu}
  
\author{Licheng Liu}
\affiliation{%
  \institution{University of Minnesota - Twin Cities}
  \city{Minneapolis}
  \state{MN}
  \country{USA}}
\email{lichengl@umn.edu}

\author{Qing Zhu}
\affiliation{%
  \institution{Lawrence Berkeley National Lab}
  \city{Berkeley}
  \state{CA}
  \country{USA}}
\email{QZhu@lbl.gov}

\author{Runlong Yu}
\affiliation{%
  \institution{University of Pittsburgh}
  \city{Pittsburgh}
  \state{PA}
  \country{USA}}
  \email{ruy59@pitt.edu}

\author{Zhenong Jin}
\affiliation{%
  \institution{University of Minnesota - Twin Cities}
  \city{Minneapolis}
  \state{MN}
  \country{USA}}
\email{jinzn@umn.edu}

\author{Yiqun Xie}
\affiliation{%
  \institution{University of Maryland}
  \city{College Park}
  \state{MA}
  \country{USA}}
\email{xie@umd.edu}

\author{Xiaowei Jia}
\affiliation{%
  \institution{University of Pittsburgh}
  \city{Pittsburgh}
  \state{PA}
  \country{USA}}
  \email{xiaowei@pitt.edu}


\begin{abstract}

Evaluating ecological time series  is critical for benchmarking model performance in many important applications, including 
predicting greenhouse gas fluxes, capturing carbon-nitrogen dynamics, and monitoring hydrological cycles. 
Traditional numerical metrics (e.g., R-squared, root mean square error) have been widely used  to quantify the  similarity between modeled and observed ecosystem variables, but they 
often fail to capture domain-specific temporal patterns critical to 
ecological processes. 
As a result, these methods are often accompanied by expert visual inspection, which requires substantial human labor and limits the applicability to large-scale evaluation. 
To address these challenges, we propose a novel framework that integrates metric learning with large language model (LLM)-based natural language policy extraction to develop interpretable evaluation criteria. 
The proposed method processes pairwise  annotations and implements a policy optimization mechanism to generate and  combine different assessment metrics. 
The results obtained on multiple datasets for evaluating 
the predictions of crop gross primary production and carbon dioxide flux    have confirmed  the effectiveness of the proposed method in capturing target assessment preferences, including both synthetically generated and expert-annotated model comparisons. 
The proposed framework  bridges the gap between numerical metrics and expert knowledge while providing interpretable evaluation policies that accommodate the diverse needs of different ecosystem modeling studies.

\end{abstract}

\maketitle

\section{Introduction}

Ecological models in environmental science are  increasingly used by government, resource managers, and companies to inform critical decision-making across various domains, including  precision farming, conservation planning, disaster management, insurance estimation, and  climate policy development.  
These models use either physics-based approaches (e.g., process-based models~\cite{fatichi2016overview,zhou2021quantifying,hipsey2019general,cuddington2013process,markstrom2015prms}) or data-driven approaches (e.g., machine learning models~\cite{liu2024knowledge,nguyen2023climax,reichstein2019deep,jia2021physics_pgrgrn,schmude2409prithvi}) to  simulate  dynamics of key variables within the target ecosystems. 
Given their influence on policy and practical decision-making, 
effective assessment of ecological models  becomes essential for ensuring the reliability of  the model outputs, and can further help mitigate 
critical societal risks such as  misguided environmental policies and inefficient resource allocation. 


A common approach for assessing ecological models is the visual inspection of time series predictions. Experts analyze graphical representations to identify patterns, trends, and anomalies. This method provides a nuanced understanding of model performance, capturing complex ecosystem dynamics. However, visual assessments are inherently subjective, and interpretations can be inconsistent among evaluators. 
Moreover, as ecological studies expand in scale, relying solely on visual assessments becomes increasingly impractical due to the extensive time and effort required. 

\begin{figure}[!t]
\centering
\includegraphics[width=0.88\linewidth]{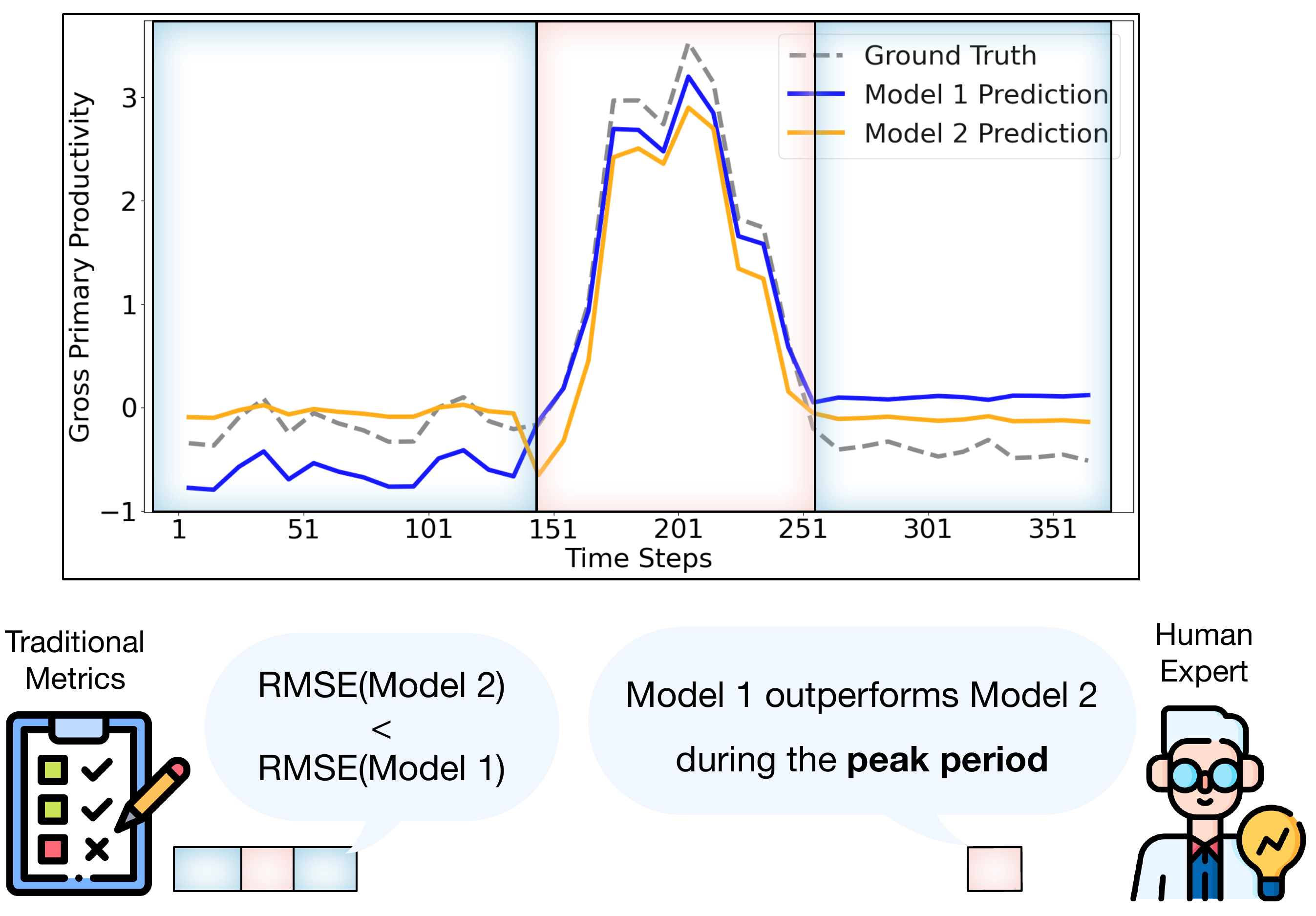}
\vspace{-0.4cm}
\caption{Illustrative example showing the inconsistency in model ranking by traditional metric (e.g., RMSE) and human expert (e.g., emphasizing peak period alignment). }
\label{fig:toy_exp}
\end{figure}

On the other hand, traditional quantitative metrics, such as root mean squared error (RMSE), R-squared, and Pearson correlation,  have also been widely used to compare model outputs with true observations.  
Each of these metrics focuses on the assessment of a specific aspect of time series, e.g., the overall magnitude match or the strength of linear association between model predictions and true labels. 
However, these metrics are unable to capture the complex nature of ecosystem dynamics and reflect  
desired ecosystem behaviors, 
such as the  phase alignment of seasonal cycles, response to extreme weather events, and derivative relationships between interacting variables~\cite{collier2018international,gauch2023defense}. 
Moreover, existing metrics often  collapse the evaluation of long temporal data into a single overall performance score, without explicitly distinguishing between different time periods. 
Figure~\ref{fig:toy_exp} illustrates an example in which Model 1 has higher RMSE than Model 2 while human expert considers Model 1 to be superior due to its better alignment with ground truth observations in the peak period. 
Finally, traditional evaluation metrics may not adequately serve the diverse needs of different studies. In particular, different scientific communities prioritize distinct temporal patterns in their evaluations, e.g.,  agronomists emphasize growing season dynamics of crops while climatologists focus on crops' interannual variability. 

Existing efforts address these issues by combining multiple quantitative measures including both traditional evaluation metrics and domain-specific measures~\cite{collier2018international,konidari2007multi}. These composite metrics are often designed for a specific task and thus rely on preset weighted combination and cannot adapt to varying expert requirements. 
Existing large language models (LLMs)-based evaluation methods~\cite{lin2023llm,wang2023pandalm,zhu2024adaptive,panickssery2024llm} face the similar limitation of adapting to different assessment priorities and are also   focused on  evaluating  text or image generation tasks. 
On other hand, prior work explored building learnable similarity measures for time series data \citep{kazemi2019time2vec,yue2022ts2vec}, but these ``black box'' metrics provide limited interpretability about the time series characteristics being captured. 



In this paper, 
we introduce the Adaptive Policy Evaluation Framework (APEF), a novel LLM-based methodology that adaptively formalizes expert visual assessment patterns   into  interpretable evaluation policies for a target evaluation task. 
By bridging the gap between quantitative metrics and qualitative expert judgment, APEF advances the rigor and transparency of model evaluation in ecological modeling. 
Developed in collaboration with scientists in different ecological modeling domains, 
APEF provides a foundation for developing assessment systems where different evaluation policies can be standardized for specific contexts while maintaining adaptability to emerging scientific priorities. 
The evaluation process  is also highly interpretable as it offers details about how specific temporal patterns influence model comparisons. 
APEF has the potential to support  reliable decision-making in many scientific domains, such as sustainable agriculture and climate change mitigation.



In particular, the proposed APEF 
pursues the innovation of using an LLM to automatically extract evaluation policies from expert judgements.  
Instead of having the LLM learn the policy from scratch, we propose to leverage the guidance from a base metric to capture representative  characteristics in ecological time series. The base metric combines multiple domain-specific metrics with a learnable weighted combination and is updated in parallel with an LLM-based policy generator. 
The LLM-based policy generator iteratively 
synthesizes human-readable evaluation policies by leveraging the optimization history for the base metric and expert judgments.  
This approach also enables expanding the assessment priorities beyond those covered in the base metric and documenting how specific assessment aspects contribute to the final assessment outcome.






We validate the framework through three complementary studies. Quantitative benchmarking against expert rankings of GPP and CO$_2$ flux predictions demonstrates  better or comparable agreement than diverse baseline metrics, significantly outperforming traditional approaches. Moreover, APEF is shown to be able to extract interpretable policies that reflect key assessment priorities. 


\section{Related Work}
\subsection{Evaluation of Ecosystem Modeling}

Evaluation of ecological models is important as these models are frequently used for informing decision-making and advancing scientific understanding of ecological processes. 
Such evaluation is highly challenging as it needs to consider desired data characteristics 
for target studies and target communities. 
For example, when modeling carbon budget in agricultural ecosystems, agronomists focus on 
growing season alignment due to its impact on crop productivity and resource efficiency 
while climatologists emphasize annual carbon budgets \citep{IPCCAgriculture2022}. Another common focus of evaluation  is on  extreme event responses (droughts, floods), which require non-linear scoring that standard metrics fail to capture \citep{cloke2008evaluating,easterling2016detection}.

Current evaluation of ecological systems  relies on either visual assessment or pre-defined metrics and rules~\cite{gauch2023defense,garrett2024validating}. Visual assessment requires substantial human labor  while pre-defined  metrics can only capture a specific aspect of data properties. 
Prior work has also explored combining different evaluation metrics~\cite{collier2018international,konidari2007multi}, but they often rely on 
fixed weights or manual tuning, which fails to capture the nuanced requirements of specific evaluation tasks. 
ML-based methods~\cite{kazemi2019time2vec,yue2022ts2vec} provide the potential to automatically  encode  complex time series patterns  and efficiently perform the similarity measurement between samples. Model ensemble can also be used to estimate model uncertainty~\cite{flato2014evaluation}. 
However, deploying the evaluation process in real systems often  demands interpretable metrics or rules traceable to known ecological principles – a requirement unmet by black-box learned metrics. 

\subsection{LLMs in Scientific Evaluation}
Large language models have revolutionized scientific workflows through their ability to parse domain literature and perform diverse text-related tasks \citep{Boiko2023emergent}. In particular, LLMs have shown their ability to process complex, unstructured data samples and provide interpretable assessment criteria for evaluating text and image data~\cite{lin2023llm,wang2023pandalm,zhu2024adaptive,panickssery2024llm}. Prior work also reports that LLMs can produce text evaluation similar to human judgements~\cite{chiang2023can}. 
Different from these tasks, the evaluation of ecological modeling tasks requires the reasoning of whether each model captures key temporal dynamics related to underlying ecological processes. 
Our work advances this frontier through 
the innovation of leveraging LLM-based for learning and combining metrics for modeling time series data based on expert feedback. 

\section{Problem Formulation and Base Metric}\label{sec:method}

\begin{figure*}[t]
\centering
\includegraphics[width=0.85\textwidth]{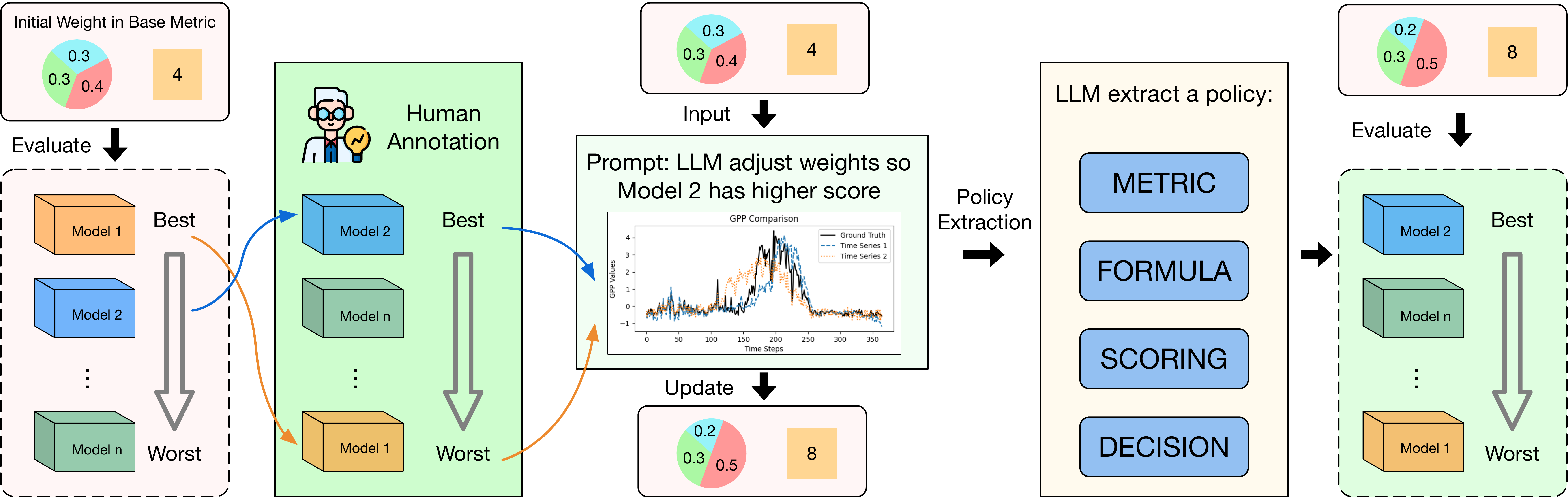}
\caption{Overall flow of APEF. LLM is first used to adjust weights in the base metric to align with expert annotation. The weight adjustment records are then fed to the LLM to extract  interpretable evaluation policies.   }
\label{fig:framework}
\vspace{-.1in}
\end{figure*}

We aim to develop evaluation policies for the time series of key ecological variables through iterative refinement guided by LLMs. The proposed APEF combines a modular base metric with a policy extraction mechanism, which adapts evaluation criteria based on expert preferences.  
Specifically, given the model-predicted time series for $N$ instances (e.g., locations or years)  $\mathcal{P} = \{P^{1}, P^{2}, ..., P^{N}\}$, where each time series $P^{i}$ represents predictions over $T$ timesteps, $P^{i} = \{p_1^{i}, p_2^{i}, 
..., p_T^{i}\}$,  APEF first computes a base similarity score by comparing the model predictions $\mathcal{P}$ with 
corresponding ground truth observations $\mathcal{Y} = \{y^{1}, y^{2}, ..., y^{N}\}$. 
The base score employs 
configurable components related to temporal patterns, peak behaviors, and derivative patterns (e.g., slope and curvature). The LLM-based weight optimizer 
then adjusts these component weights to align with expert annotations, and feed the results to the policy extraction process. The overall flow of APEF is shown in Figure~\ref{fig:framework}. 

Here we consider pairwise annotations of expert preference, as this is usually more labor efficient compared to directly scoring each individual time series or ranking multiple time series. Without loss of generality, we assume access to a set of pairwise samples $\{(P^{i,(a)}, P^{i,(b)})\}_{i,a,b}$, where each pair consists of two time-series samples $(a)$ and $(b)$ (e.g., predicted by two different models) over a specific data instance $i$, and the series $P^{i,(a)}$ has better alignment with the observations $Y^i$ than $P^{i,(b)}$  according to the expert annotation.    
We create pairwise samples to be labeled from a set of time series data, ensuring that the overall ranking of all samples can be reconstructed from the obtained pairwise preferences.

For each weight optimization step $d$,  
APEF employs a policy extraction mechanism to derive interpretable evaluation rules $\pi_d$.  
The policy optimizer  validates the obtained policies against a validation set $\mathcal{V}$ to ensure 
consistent evaluation performance and adaptability to different evaluation contexts.


\subsection{Base Metric}\label{sec:base-metric}
Ecological time series, 
such as plant gross primary productivity and  greenhouse gas (GHG) flux, exhibit complex temporal patterns driven by  underlying ecological processes. Traditional evaluation metrics like RMSE or correlation coefficients often fail to capture many important domain-specific characteristics, as they treat all time points equally and ignore critical temporal dependencies. We develop a base metric system that decomposes time series similarity into three key components: temporal segmentation, peak behavior analysis, and derivative relationships, which jointly capture the essential patterns that domain experts consider when evaluating ecological model predictions~\cite{collier2018international,gauch2023defense,flato2014evaluation}. 

The proposed base metric is built on the recognition that ecological time series contain periods of varying importance. 
The design of our metric is based on the observation that ecological time series contain periods of varying importance.
For instance, growing seasons of plants typically exhibit more dynamic changes and better reflect the impact of external drivers compared to the dormant periods,  thus requiring more careful evaluation. To address this, we implement a temporal segmentation approach that automatically identifies these critical periods through rise-and-fall pattern detection. Given the time series of a target variable $P^{i} = \{p_1^{i}, p_2^{i}, ..., p_T^{i}\}$, the algorithm analyzes first-order differences $\Delta P^{i}_t = p^{i}_{t+1} - p^{i}_t, t \in \{1, \ldots, T-1\}$ to detect sustained directional changes. 
We then compare these differences with pre-defined thresholds, as follows: 
\begin{equation}
\small
    \begin{aligned}
        \text{RisingPeriod} &= \{ t \mid \Delta P^{i}_t > \theta_\text{rise}, \,\,\, t \in \{1, \ldots, T-1\} \} \\
        \text{FallingPeriod} &= \{ t \mid \Delta P^{i}_t < \theta_\text{fall}, \,\,\, t \in \{1, \ldots, T-1\} \} 
    \end{aligned}
\end{equation}
The thresholds $\theta_\text{rise} = 0.01$ and $\theta_\text{fall} = -0.01$ are empirically determined from our tests. 
To filter noise and independent data oscillations, we keep only the time steps in RisingPeriod or FallPeriod if they extends consecutively for at least $\theta_\text{period}$ timesteps. 
The hyper-parameter $\theta_\text{period}$ is set as 5 in our tests. 
The segmentation method aims to identify a rising period followed by a falling period, with the interval between them serving as the basis for peak analysis. 
In the following discussion, we assume each time series has only one rise-fall pattern occurrence. For longer time series with multiple occurences of rise-fall patterns, we can apply this segmentation method to create shorter sequences. 

To quantify the similarity score for peak values, we employ a soft matching approach for peak comparison that allows flexible temporal alignment within a tolerance window. Given a model prediction $P^{i} = \{p_1^{i}, p_2^{i}, ..., p_T^{i}\}$ and ground truth $Y^{i} = \{y_1^{i}, y_2^{i}, ..., y_T^{i}\}$, we define the timestep of peak, $\text{Peaks}(P^i)$, for the time series $P^i$ to be the timesteps of all the local maxima. This is formally expressed as follows:
\begin{equation}
\small
    \begin{aligned}
        \text{Peaks}(P^{i}) &= \{ t \mid p_t^{i} > p_{t-1}^{i} \text{ and } p_t^{i} > p_{t+1}^{i},  t \in \{2, \ldots, T-1\} \} \\
        \text{PeakValues}(P^{i}) &= \{ p_t^i \mid t \in  \text{Peaks}(P^{i}) \} \\
    \end{aligned}
\end{equation}

Now given $\text{Peaks}(P^{i}), \text{Peaks}(Y^{i})$, we calculate the similarity score for peak alignment. Formally, we define $S_\text{Peak}(P^{i}, Y^{i})$ combining  both peak time step alignment and peak value matching, as follows:
\begin{equation}
\small
    \begin{aligned}
        S_{\text{PeakX}}(P^{i}, Y^{i}) &= \sum_{t_p \in \text{Peaks}(P^{i})} 
        \sum_{t_y \in \text{Peaks}(Y^{i})} 
        \left( 1 - \frac{|t_p - t_y|}{\text{tolerance}} \right) \\
        S_{\text{PeakY}}(P^{i}, Y^{i}) &= \sum_{v_p \in \text{PeakValues}(P^{i})} 
        \sum_{v_y \in \text{PeakValues}(Y^{i})} 
        \left( 1 - \frac{|v_p - v_y|}{\max(v_p, v_y)} \right) \\
        S_{\text{Peak}}(P^{i}, Y^{i}) &= (1 - w_{\text{amp}}) * S_{\text{PeakX}}(P^{i}, Y^{i}) 
        + w_{\text{amp}} * S_{\text{PeakY}}(P^{i}, Y^{i})
    \end{aligned}
\end{equation}
where $ S_\text{PeakX}$ measures the  alignment of peak time steps and $S_\text{PeakY}$ captures amplitude similarity between peaks. Tolerance and $w_{amp}$ are parameters adjustable in the weight optimization process detailed in Section~\ref{sec:wgt_opt}.  
This soft-matching formulation accommodates phase shifts while maintaining sensitivity to magnitude differences. These are crucial considerations in ecological modeling where both timing and intensity of events matter. 

The proposed metric further incorporates another component to capture system dynamics through derivative analysis. Standard evaluation metrics, such as RMSE,  might show high similarity (or low errors) between two series with matching values even they exhibit  different rates of changes.  
In contrast, our approach explicitly considers slope and curvature. These derivative terms can reflect important ecosystem characteristics, such as vegetation growth and extreme events, while also representing important properties in many governing differential equations of  ecosystems~\citep{hipsey_general_2019,katzin2022process,zhou2021quantifying}. 
Specifically, we consider both  $P^{i}$ and corresponding observation series $Y^{i}$  as a function of timestep $t$, so we have $P^{i}(t)$ and $Y^{i}(t)$. Then we measure the slope and curvature of each series (e.g., for $P^{i}(t)$) as 
$\quad \frac{dP^{i}(t)}{dt}$ and $\quad \frac{d^2P^{i}(t)}{dt^2}$. The derivative-based distance score between $P^{i}$ and $Y^{i}$ is computed as follows: 
\begin{equation}
\small
    \begin{aligned}
        S_{\text{slope}}(Y^i, P^i) &=\frac{1}{T} \sum_{t=1}^{T} (\frac{dP^{i}(t)}{dt} - \frac{dY^{i}(t)}{dt})^2  \\
        S_{\text{curv}}(Y^i, P^i) &=\frac{1}{T} \sum_{t=1}^{T} (\frac{d^2P^{i}(t)}{dt^2} - \frac{d^2Y^{i}(t)}{dt^2})^2  \\
        S_{\text{Deriv}}(Y^i, P^i) &=  w_\text{der} * (S_{\text{slope}}(Y^i, P^i) +  S_{\text{curv}}(Y^i, P^i)) 
    \end{aligned}     
\end{equation}
where $w_\text{der}$ is a scaling factor and can be adjusted 
to reflect the varying contribution of derivative-based  difference to the overall alignment between model prediction and ground truth. 
This multi-order analysis enables capturing both immediate changes and overall trends, which is  
essential for evaluating model responses to environmental perturbations. 

Finally, we compute the base metric score combining these components with period-specific weighting. 
Given RisingPeriod and FallingPeriod of the observations $Y^i$, we define the $t_\text{st}$ as the starting time step 
in RisingPeriod, $t_\text{ed}$ 
as the ending timestep in FallingPeriod. 
For each time series ($P^i$ or $Y^i$), we segment it into three parts, $P^i_{\text{before}} = \{ p_t^i \mid t < t_\text{st} \}$, $P^i_{\text{in}} = \{p_t^i \mid t_\text{st} < t < t_\text{ed}  \}$, and $P^i_{\text{after}} = \{ p_t^i \mid t > t_\text{ed} \}$. We  compute alignment scores separately for each part and then combine them to get the final base metric score, as follows: 
\begin{equation}
\small
   \begin{aligned}
       S_{\text{in}}(Y^i, P^i) &= S_{\text{Peak}}(P^{i}_{\text{in}}, Y^{i}_{\text{in}}) 
       + S_{\text{Deriv}}(P^{i}_{\text{in}}, Y^{i}_{\text{in}}) \\
       S_{\text{before}}(Y^i, P^i) &= 
       S_{\text{Peak}}(P^{i}_{\text{before}}, Y^{i}_{\text{before}})  +  
       S_{\text{Deriv}}(P^{i}_{\text{before}}, Y^{i}_{\text{before}})\\
       S_{\text{after}}(Y^i, P^i) &=  
       S_{\text{Peak}}(P^{i}_{\text{after}}, Y^{i}_{\text{after}}) 
        +   
       S_{\text{Deriv}}(P^{i}_{\text{after}}, Y^{i}_{\text{after}})\\
       S(Y^i, P^i) &= w_{\text{peak}} S_{\text{before}}(Y^i, P^i) \\
       &+  \frac{1 - w_\text{peak}}{2}
       (S_\text{in}(Y^i, P^i) + S_{\text{after}}(Y^i, P^i)) 
   \end{aligned}
   \label{eq:base_metric}
\end{equation}




This base metric score can also be converted to a similarity measure as $S(Y^i,P^i)=\frac{1}{1+S(Y^i,P^i)}$. This formulation allows flexible emphasis on different temporal segments while maintaining a normalized score range, facilitating consistent model comparison across different scenarios and time scales.

\section{LLM-based Policy Extraction}

We now describe how to use the LLM to help generate the policy for evaluating time series outputs. Rather than having LLM learn complex policies from scratch, we use the base metric to provide guidance to the policy learning process. Specifically, we first use LLM to optimize each weight parameter in the base metric to align with the expert preferences. Then we feed the obtained results to a parallel policy extraction module to synthesize the evaluation rules.

\subsection{Weight Optimization in Base Metric}
\label{sec:wgt_opt}

The base metric can prioritize different assessment aspects (e.g., peak alignment, curvature match) by adjusting its weight parameters. Here we use $\mathbf{w}$ to represent all the adjustable parameters in the base metric. 
Optimizing these weights to align with expert preferences is challenging due to (i) the highly non-linear relationship between weight parameters and the alignment with pairwise preferences,  
and (ii) the desired constraints over the  weights 
to maintain their interpretability. For example, we may expect each weight to be positive, with their values normalized over all the weights.

We address this issue through an LLM-based weight optimization mechanism that dynamically adjusts metric parameters based on the predicted scores and pairwise expert preferences. 
Specifically, given a pair of time series samples $(P^{i,(a)}, P^{i,(b)})$ for the same instance $i$, assuming expert prefers sample $(a)$ over sample $(b)$,  
the optimizer aims to adjust {weights $\mathbf{w}$} such that the estimated base metric scores (Eq.~\ref{eq:base_metric}) for these two samples align with the expert preferences:
\begin{equation}
    S(Y^i, P^{i,(a)}) > S(Y^i, P^{i,(b)})
    \label{eq:pair_preference}
\end{equation}


Here we employ an LLM-based approach to 
iteratively update weight parameters $\mathbf{w}$ 
by leveraging a given pair of expert-annotated time series samples  and   the optimization history in a structured prompt.  At each iteration $d$, the LLM  takes the input of the current estimated weights $\mathbf{w}_d$, the optimization history $\mathcal{H}_d$ until the current iteration, the given pairwise sample and corresponding observations $P^{i,(a,b)} = \{P^{i,(a)}, P^{i,(b)}, Y^i\}$ , and the optimization constraints $\mathcal{C}$, as follows:
\begin{equation}
    \mathbf{w}_{d+1} = \text{LLM}(\mathbf{w}_{d}, \mathcal{H}_{d}, P^{i,(a,b)}, \mathcal{C})
\end{equation}

Here the  optimization history $\mathcal{H}_{d} = \{(\textbf{w}_{d'}, \textbf{P}_{d'}, \rho_{d'})\}_{{d'}=1}^{d}$ represents the optimization history over $d$ iterations,  
where $\textbf{w}_d$, $\textbf{P}_d$, and $\rho_d$ represent the updated weights, the pairwise time series samples considered in the iteration $d$, and the overall correlations obtained from the training samples at each iteration $d$, respectively. Here the correlation is measured by comparing  the rankings of all the time series in the training set,  as determined using the base metric (Eq.~\ref{eq:base_metric}) with the current weight at iteration $d$, against the rankings derived from expert annotations. 

To ensure stable weight adjustments over iterations, we incorporate several practical constraints in $\mathcal{C}$, as follows:
\begin{itemize}
    \item Weight bounds: each parameter in $\textbf{w} \in [0.1, 1.0]$.  
    \item Smoothness: $|\mathbf{w}_{t+1} - \mathbf{w}_t| \leq \delta$. 
    \item Normalization: $\sum_{w_i\in \textbf{w}} w_i = 1$ for the weights. 
\end{itemize}

We include an example of optimization history and the entire LLM input in Appendix \ref{sec:appenix:WO}. 
After each update iteration, the LLM response is parsed to extract new weight values. 
This approach offers  threefold benefits. 
First, it can leverage patterns in historical performance to make informed adjustments. Second, it maintains interpretability by providing reasoning for each weight change. Third, it can adapt to different evaluation contexts by considering the specific characteristics of the time series being compared.

    
    
    



\begin{figure}[!t]
\centering
\includegraphics[width=0.48\textwidth]{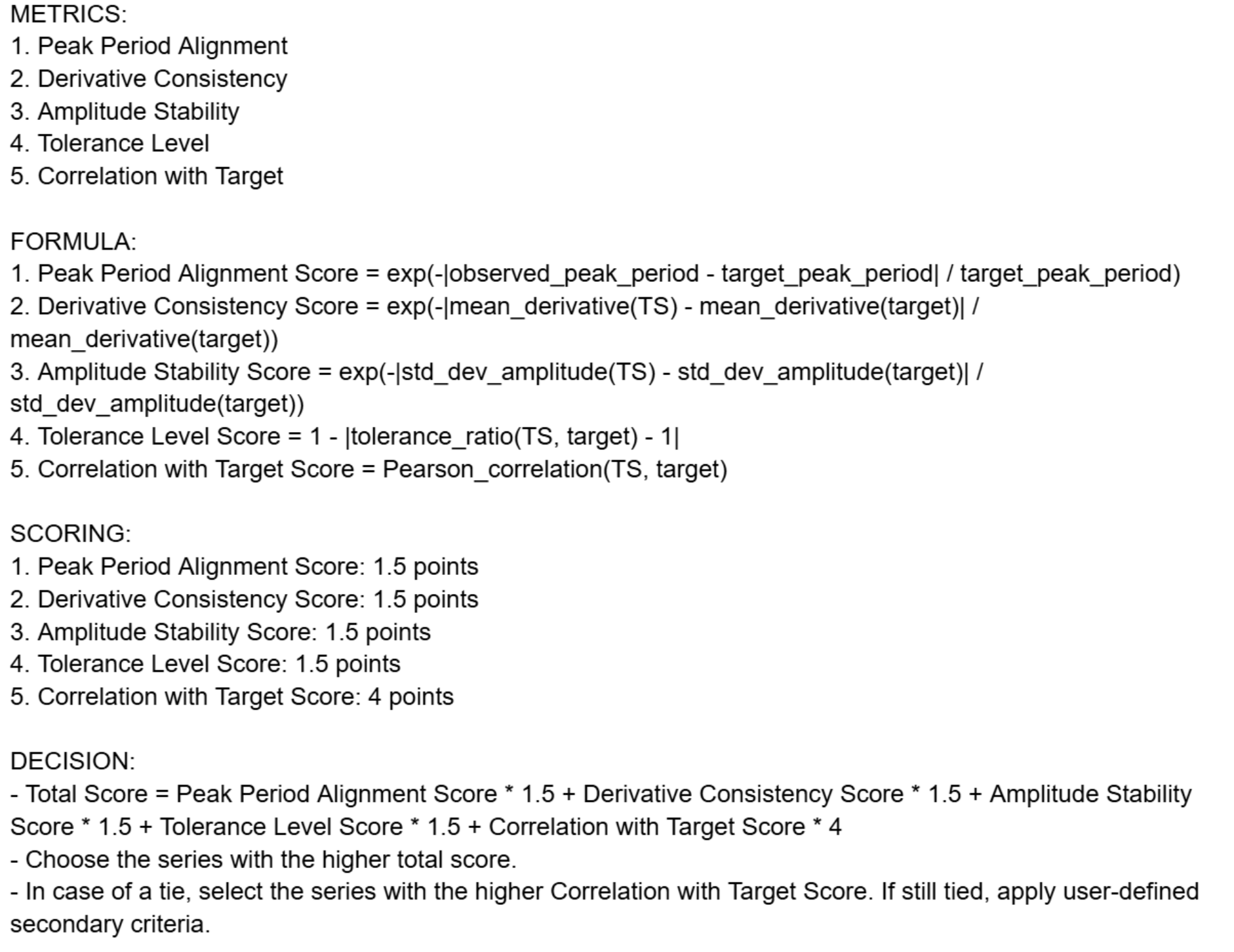}
\vspace{-.1in}
\caption{An example of policy extracted by APEF. 
}
\label{fig:policy_example}
\end{figure}
\subsection{Policy Extraction }

While the weight optimization process can adjust the parameters in the base score,  
it does not provide the underlying rationale for evaluation preferences           and also cannot capture assessment aspects that are not covered by the base metric. We address these limitations through a policy extraction mechanism that translates the optimization history into interpretable evaluation rules. This approach leverages the outcome of the weight optimization process, while also continuously operating in parallel  to refine natural language policies for explaining the relationship between time series characteristics and target preferences.

At each iteration $d$, given the optimization history $\mathcal{H}_d$, 
the policy extractor generates a structured evaluation policy $\pi_d$ consisting of four key components, as  
\begin{equation}
    \pi_d = \{\mathcal{M}_d, \mathcal{F}_d, \mathcal{S}_d, \mathcal{R}_d\},
\end{equation}

We describe each component below. 
\begin{itemize}
    \item \textbf{Metrics $\mathcal{M}_D$}: Name and interpretation of assessment metrics. Each assessment metric measures the similarity between two time series on a certain aspect, e.g., peak alignment, derivative similarity, as follows:
    \begin{equation}
        \mathcal{M}_d = \{m_1: \text{peak alignment}, m_2: \text{seasonal patterns}, ...\}
    \end{equation}

    It is noteworthy that these assessment metrics are automatically extracted by LLM and could be different from the proximity terms considered in the base metric.
    
    \item \textbf{Formula}: Notably, assessment metrics can be interpreted differently, especially when a metric is not widely known. Thus, we ask LLMs to provide a mathematical formulation for each metric $m \in \mathcal{M}_d$. 
    \begin{equation}
    \begin{aligned}
        \mathcal{F}_d = &\{\text{peak tolerance}\,\,\, m_1= f_1(P^i, Y^i), \\
        &\text{amplitude ratio}\,\,\, m_2=f_2(P^i, Y^i), ...,\}
    \end{aligned}
    \label{eq:formula}
    \end{equation}
    
    The incorporation of  this component helps reduce the uncertainty of the obtained policies. We notice that given the same set of samples, LLMs may produce  different evaluation results over multiple runs if the metrics are not explicitly defined with mathematical formulations.  

    \item \textbf{Scoring}: This component quantifies the contribution for each assessment metric. Specifically, 
    we  assign a positive score  $p$ for each metric $m$, as 
    \begin{equation}
        \mathcal{S}_d = \{(m_1, p_1), (m_2, p_2), ...\}, \,\text{ s.t. }\, \sum p_i = K, 
    \end{equation}
    where the constraint imposed by $K$ is to enforce the competition between different metrics. We set $K=10$ in our tests.  
    The scores $p$ are used to scale the original value obtained from its formula in Eq.~\ref{eq:formula}, which will then be combined for final assessment measurement.

    \item \textbf{Decision Rule}: 
    The final evaluation rule consists of two levels: (i) It first creates an overall assessment score by aggregating the scores from different assessment metrics using a LLM-generated mathematical form, as
    \begin{equation}
    f_\text{aggr}(m_1 p_1, m_2 p_2, ...).
    \end{equation}
    (ii) Additionally, in cases where multiple series (e.g., $P^{i,(a)}$ and $P^{i,(a)}$) have the same overall score, 
    the LLM generates additional rules by referring to the original metric values.    
\end{itemize}

This structured policy 
combines multiple metrics and different numerical operations, which facilitates capturing the complex assessment interpretation of domain experts. 
Additionally, the policies provide clear rationales for evaluation decisions, enabling domain experts to understand and verify the learned evaluation criteria.


\subsection{Policy Optimization}

Synthesizing policy from history data is challenging due to the need to account for diverse time series characteristics as assessment metrics. The adjustment of these metrics and their associated scores also need to be reconciled.


We iteratively update the policy by incorporating the optimization history $\mathcal{H}_d$, a pairwise time series sample, and the current policy. 
To ensure a smooth policy update while enhancing generalizability, we propose a two-stage policy update approach:  

\textbf{Update of assessment metrics $\mathcal{M}_D$: } In each iteration, in addition to updating other components of the policy, we permit the addition or removal of at most one assessment metric $m$ along with its associated  score $p$. When introducing a new assessment metric $m$, we tentatively use it in isolation to compare each pair of time series in the expert-annotated training data. The new assessment metric is retained only if the obtained comparisons align with expert annotation for a substantial portion of the training pairs.

    
    

\textbf{Policy-level validation: }
After updating the policy, we validate the new policy using a separate validation set, which is derived from a subset of the original training pairwise data. Specifically, we  measure the performance of the updated policy $\pi_{d+1}$ on the validation set and compare it with the policy $\pi_{d}$ obtained in the previous iteration. Additionally, due to the complexity of the policy, we notice that the LLM may produce different evaluation outcomes in different runs even though we explicitly specify the calculation of each assessment metric. Hence, we will have multiple runs over the validation set using the new policy. The new policy will be retained only when a certain portion of these runs produce better validation performance. This process is expressed as follows: 
\begin{equation}
    \pi_\text{new} = \begin{cases}
        \pi_{d+1} & \text{if } \rho_{\text{val}}(\pi_{d+1}) > \rho_{\text{val}}(\pi_d) \,\,\text{for\, $\theta$\, LLM runs},
        \\
        \pi_d & \text{otherwise},
    \end{cases}
\end{equation}
where the threshold $\theta$ is set to be 70\% in our tests. 
To ensure the optimizer evaluates policies over a comprehensive set of scenarios, we maintain detailed logs of policy updates, validation performance, and component success rates. This information is crucial for analyzing the policy optimization process and understanding the characteristics of successful evaluation criteria.

\subsection{Extension to Evaluating Multi-variate Series}
Ecological models often simultaneously simulate multiple key variables involved in the target ecosystem, and many variables can be highly correlated with each other. If an ecological  model fails to preserve these underlying relationships, it lacks physical consistency and becomes less reliable.  
Therefore, evaluation should go beyond assessing the alignment of each predicted series with its corresponding observations and also consider the interdependencies among predicted time series for related variables. 

We extend the proposed APEF to simultaneously evaluate two time series of related variables, e.g., crop gross primary production and CO$_2$ flux. We can  represent the input pairwise sample as $(\{P^{i,(a)}, Q^{i,(a)}\}$, $\{P^{i,(b)}, Q^{i,(b)}\})$, i.e., predictions of two variables $P$ and $Q$  by two models $(a)$ and $(b)$.   
The key idea is to harness the ability of the LLM to formulate evaluation policies that incorporate association constraints between the variables. To our knowledge, APEF is also the first automatic evaluation method that takes into account the relationship between multiple time series.  

During the weight optimization process (Section~\ref{sec:wgt_opt}), we need to aggregate the base metric scores obtained for the given two variables to estimate the overall combined score. Specifically, the combined score is computed as the average of the individual base metric scores, scaled by their correlation difference, as follows: 
\begin{equation}
    \small
    \begin{aligned}
    \tilde{S}^{i,(a)} = \frac{S(Y_p^i, P^{i,(a)}) + S(Y_q^i, Q^{i,(a)})}{2} \text{Corr} (Y^i_p,Y^i_q,P^{i,(a)}, Q^{i,(a)}), \\
    \text{Corr}(Y^i_p,Y^i_q,P^{i,(a)}, Q^{i,(a)}) = 1 - \left| \text{Spear}(Y_p^i, Y_q^i) - \text{Spear}(P^{i,(a)}, Q^{i,(a)}) \right|, 
    \end{aligned}
\label{eq:score_2}
\end{equation}
where $Y_p^i$ and $Y_q^i$ represent the true observation series of the variable $P$ and $Q$ for the instance $i$, and $\text{Spear}(\cdot)$ represents the Spearman's correlation. 
We use this score to estimate the relative ordering of two samples $(a)$ and $(b)$, which is then used to update the weight parameters. It is noteworthy that the relationship expressed by Eq.~\ref{eq:score_2} may not fully capture the complex interdependence of the two variables. This score only serves as a base metric, which can be refined when APEF creates policies using the LLM. For example, the policy may define its own temporal consistency measure either by leveraging simple transformations of the correlation measure or developing a new measure from scratch. We include an example of obtained policy for two variables in Appendix \ref{sec:appendix-2f}.

\section{Evaluation}
Our evaluation is focused on the assessment predicting (1) crop gross primary production (GPP) and (2) CO$_2$ flux. The proposed APEF can also be easily applied to other prediction tasks. 
Our evaluation encompasses three distinct datasets for real-world GPP and CO$_2$ flux time series. The pairwise sample comparison is provided in different ways (e.g., by  expert annotations or pre-defined metrics) across these datasets, which are used to validate  
APEF's ability to learn different evaluation criteria. 

\subsection{Dataset Description}\label{sec:dataset}
Each dataset consists of three main components: (1) model predictions of GPP and CO$_2$ flux, (2) reference GPP and CO$_2$ flux data, (3) annotations for comparing  prediction samples. In the following, we will describe these three components for  each dataset. 


\subsubsection{Synthetic Dataset}\label{sec:synthetic-dataset}

We plan to use this synthetic dataset to evaluate if the proposed framework can recover the assessment using the base metric (described in Section \ref{sec:base-metric}) with preset weights  and produce sample preferences that correlate well with the target comparisons generated using these preset weights. 
In particular, this dataset utilizes observational time series data of CO$_2$ flux and GPP derived from 11 cropland eddy covariance (EC) flux tower sites located in major U.S. corn and soybean production regions \cite{KGML_fluxdata}. These sites, including US-Bo1, US-Bo2, US-Br1, US-Br3, US-IB1, US-KL1, US-Ne1, US-Ne2, US-Ne3, US-Ro1, and US-Ro5, span across Illinois, Iowa, Michigan, Nebraska, and Minnesota states. The dataset provides daily-scale measurements from 2000 to 2020, with each site having different operational periods ranging from 5 to 19 years. For the experiments, we used the observational values of CO$_2$ flux and GPP in year 2005 at site US-Bo1 as our ground truth.

We employed data augmentation techniques to generate synthetic model predictions from the observation time series. 
Specifically, we created synthetic prediction series data using a combination of four different augmentation methods from the TSGM (Time Series Generation Models) library~\cite{nikitin2023tsgm}: 
addition of random noise, slice and shuffle, magnitude warping, and window warping. We generated in total 20 model-predicted time series, and randomly selected 10 of them as training set, 5 of them as validation set, and 5 of them as testing set for each experiment. 





For the experiment on this dataset, we designed three sets of preset weights to evaluate different aspects of time series similarity through our base metric components: Peak Period Similarity, Derivative, and Amplitude. Each set emphasizes one component while maintaining minimal weights for the others. The preset weight combinations $(w_\text{peak}, w_\text{der}, w_\text{amp})$ are: Peak Period - $(0.8, 0.1, 0.1)$, Slope \& Curvature - $(0.1, 0.8, 0.1)$, and Amplitude - $(0.1, 0.1, 0.8)$. 
We use each preset weight combination to compute the base metric and then create the target ranking. We use Eq.~\ref{eq:score_2} to compute the score for two variables GPP+CO$_2$ and create the target ranking. 


\begin{table*}[!t]
\small
\caption{Correlation with simulated weight sets. Each column (Peak Period, Slope \& Curvature, and Amplitude) represent a preset scenario where the indicated aspect is more valued than others.}
\begin{tabular}{c|clclc|ccc|ccc}
\hline
\multicolumn{1}{l|}{} & \multicolumn{5}{c|}{Peak Period}                                                          & \multicolumn{3}{c|}{Slope \& Curvature} & \multicolumn{3}{c}{Amplitutde} \\ \hline
                      & \multicolumn{2}{c}{CO$_2$}         & \multicolumn{2}{c}{GPP}            & GPP + CO$_2$    & CO$_2$     & GPP     & GPP + CO$_2$     & CO$_2$  & GPP  & GPP + CO$_2$  \\ \hline
$R^2$                 & \multicolumn{2}{c}{0.067}          & \multicolumn{2}{c}{0.139}          & 0.600&            0.406&         0.042&                  0.527&         0.079&      0.648&               0.394\\
RMSE                  & \multicolumn{2}{c}{0.067}          & \multicolumn{2}{c}{0.139}          & 0.600&            0.406&         0.042&                  \textbf{0.588}&         0.079&      0.648&     0.406\\
MAE                   & \multicolumn{2}{c}{0.067}          & \multicolumn{2}{c}{0.127}          & 0.684&            0.406&         0.091&                  0.576&         0.188&      \textbf{0.733}&      \textbf{0.418}\\
NSE                   & \multicolumn{2}{c}{0.407}          & \multicolumn{2}{c}{0.297}          & 0.369&            0.006&         0.697&                  0.261&         0.261&      0.261&               0.248\\
TILDE-Q               & \multicolumn{2}{c}{0.067}          & \multicolumn{2}{c}{0.127}          & 0.684&            0.406&         0.091&                  0.576&         0.188&      \textbf{0.733}&               \textbf{0.418}\\
PRP-Rank              & \multicolumn{2}{c}{0.321}          & \multicolumn{2}{c}{0.067}            & 0.115&            0.115&         0.418&                  0.139&         0.152&      0.321&               0.261\\ \hline
APEF                  & \multicolumn{2}{c}{\textbf{0.452}} & \multicolumn{2}{c}{\textbf{0.300}} & \textbf{0.770}&            \textbf{0.669} &         \textbf{0.699} &    0.576 &         \textbf{0.455} &      \textbf{0.733} &              
\textbf{0.418} \\ \hline \end{tabular}
\label{tab:exp1}
\vspace{-.15in}
\end{table*}

\begin{table}[!t]
\small
\caption{Correlation with expert annotations (left) and the ILAMB scores (right). }
\begin{tabular}{c|clclcl|cc}
\hline
 & \multicolumn{6}{c|}{Annotation} & \multicolumn{2}{c}{ILAMB} \\ \hline
      & \multicolumn{2}{c}{CO$_2$}   & \multicolumn{2}{c}{GPP} & \multicolumn{2}{c|}{GPP + CO$_2$} & {CO$_2$} & GPP\\ \hline
$R^2$ & \multicolumn{2}{c}{0.167} & \multicolumn{2}{c}{0.667}  & \multicolumn{2}{c|}{0.017} & - & -            \\
RMSE& \multicolumn{2}{c}{0.167} & \multicolumn{2}{c}{0.667}  & \multicolumn{2}{c|}{0.117}   & - & -          \\
MAE& \multicolumn{2}{c}{0.167} & \multicolumn{2}{c}{0.713}  & \multicolumn{2}{c|}{0.100}          & - & -   \\
NSE& \multicolumn{2}{c}{0.367} & \multicolumn{2}{c}{0.217}  & \multicolumn{2}{c|}{0.283}     & - & -        \\
TILDE-Q & \multicolumn{2}{c}{0.167} & \multicolumn{2}{c}{0.713}  & \multicolumn{2}{c|}{0.100} & \textbf{0.857} & \textbf{0.857}            \\
PRP-Rank& \multicolumn{2}{c}{0.733} & \multicolumn{2}{c}{0.600}  & \multicolumn{2}{c|}{0.383} & 0.607 & 0.392            \\ \hline
APEF& \multicolumn{2}{c}{\textbf{0.785}} & \multicolumn{2}{c}{\textbf{0.752 }}    & \multicolumn{2}{c|}{\textbf{0.417 }}  & 0.750 & 0.821 \\ \hline      
\end{tabular}
\label{tab:exp2}
\end{table}

\subsubsection{Human Expert Alignment}\label{sec:human-dataset}

In this dataset, we conduct human expert alignment testing, where we train our framework to learn a policy that mimics expert judgment. 
We used the same observation data and synthetic prediction data as the first dataset. 
Then we collected pairwise model comparisons from three domain experts who independently evaluated the performance of different model predictions. Each expert was presented with pairs of model predictions of (1) GPP, (2) CO$_2$ flux, and (3) GPP and CO$_2$ together, and provided a comparison. Example annotation task can be found in figure \ref{fig:exp_annotation_appendix} in Appendix \ref{sec:anno}.
After getting the annotation for the pairwise comparison, we then used majority voting to create the final reference rankings. 
The inter-annotator agreement was assessed using Fleiss' kappa coefficient ($\kappa$), revealing varying levels of agreement across different aspects of model evaluation. We have $\kappa = 0.69$, $\kappa = 0.58$ and $\kappa = 0.63$ for GPP, CO$_2$, and overall evaluations respectively. For this experiment, we used the same model predictions and ground truth time series in Section \ref{sec:synthetic-dataset}. 

\subsubsection{ILAMB Score System}\label{sec:ilamb-dataset}

We also validate APEF to reproduce the model rankings in the ILAMB (International Land Model Benchmarking) scoring system~\cite{collier2018international}. 
Here the reference CO$_2$ data are from the NOAA GMD monthly flask dataset~\cite{noaa_gmd}, and the reference GPP data 
are from the FluxNet Tower eddy covariance measurements (Tier 1)~\cite{pastorello2020fluxnet2015}. 
We used model predictions from seven CMIP6 models at the BE-Bra site (year 2000) as the training set \cite{gmd-9-1937-2016}. Predictions from the same models at BE-Vie served as the validation set, while those from DE-Hai were used for testing. Using different sites ensured a sufficient sample size, as splitting predictions from only seven models would be insufficient for meaningful ranking.   
The predictions are ranked according to three core components from the ILAMB scoring system: Bias Score, RMSE Score, and Seasonal Cycle Score. 
The final ILAMB score for each prediction is computed as a weighted average of these metrics~\cite{collier2018international}. 
More details are provided in Appendix \ref{sec:appendix-ilamb}. 

\subsection{Evaluation Results}

\begin{figure} [!h]
\centering
\subfigure[Peak Period.]{ 
\includegraphics[width=0.43\textwidth]{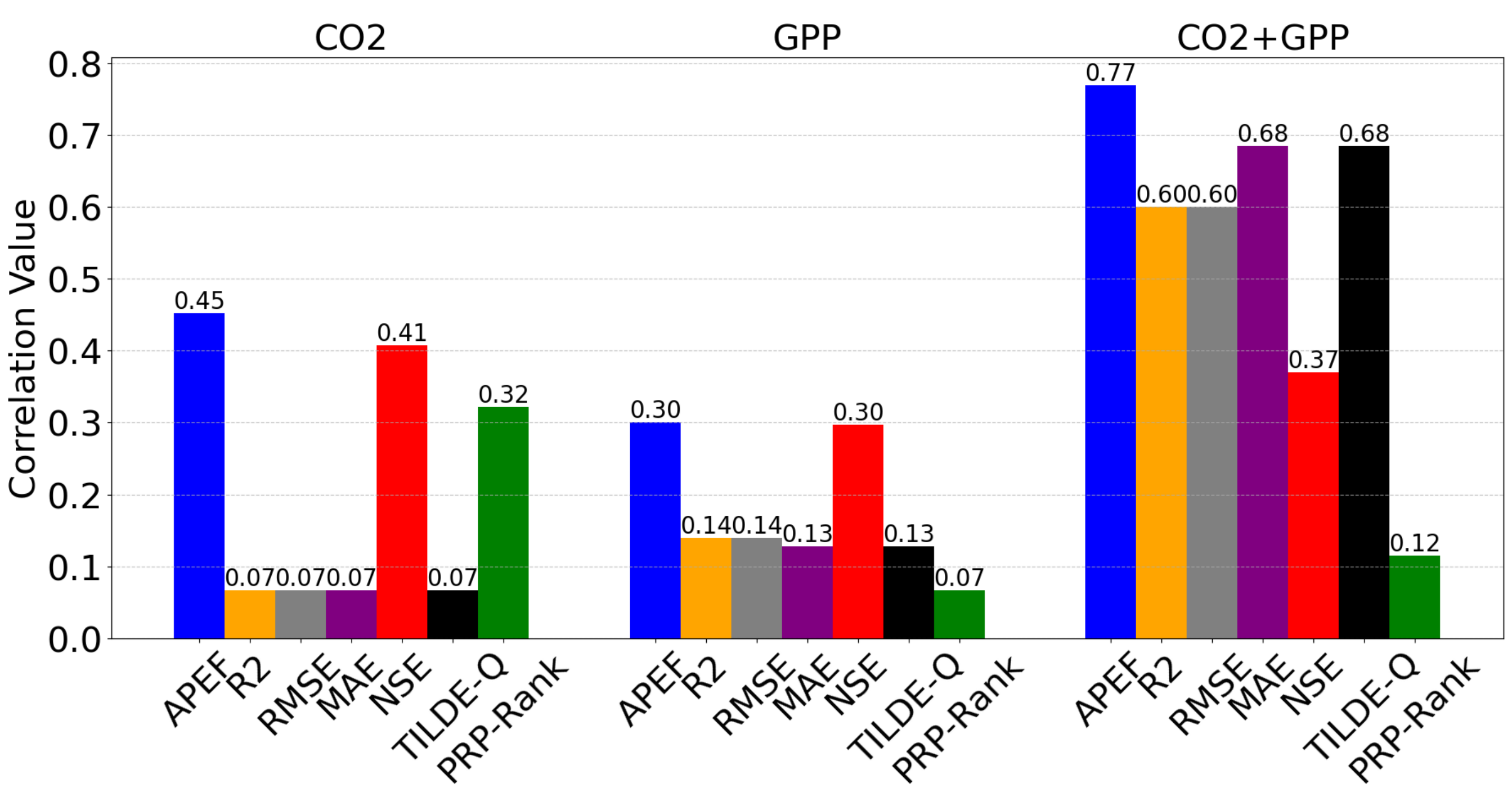}
}\vspace{-.1in}
\subfigure[Slope \& Curvature]{ 
\includegraphics[width=0.43\textwidth]{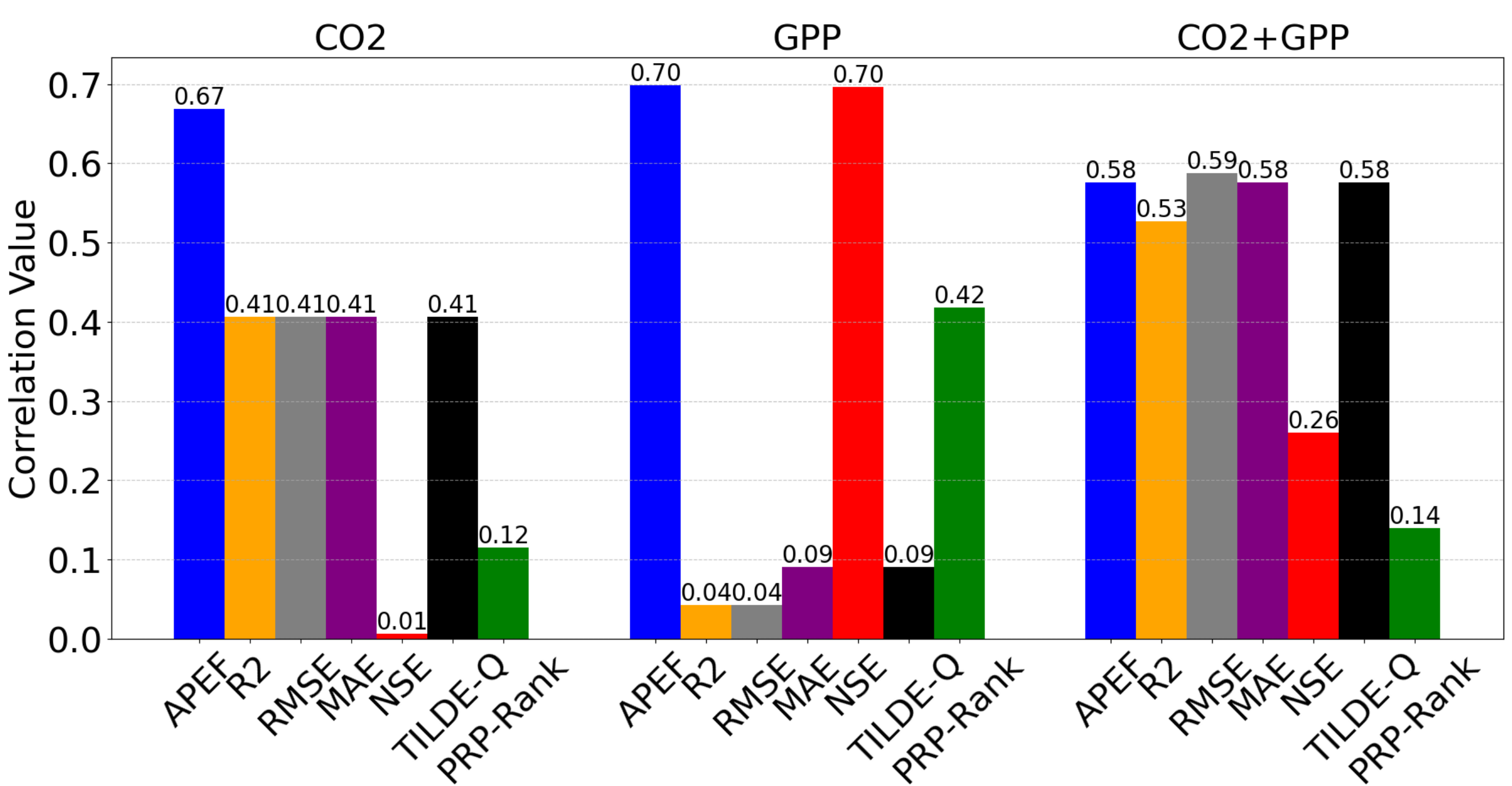}
}\vspace{-.1in}
\subfigure[Amplitude.]{ 
\includegraphics[width=0.43\textwidth]{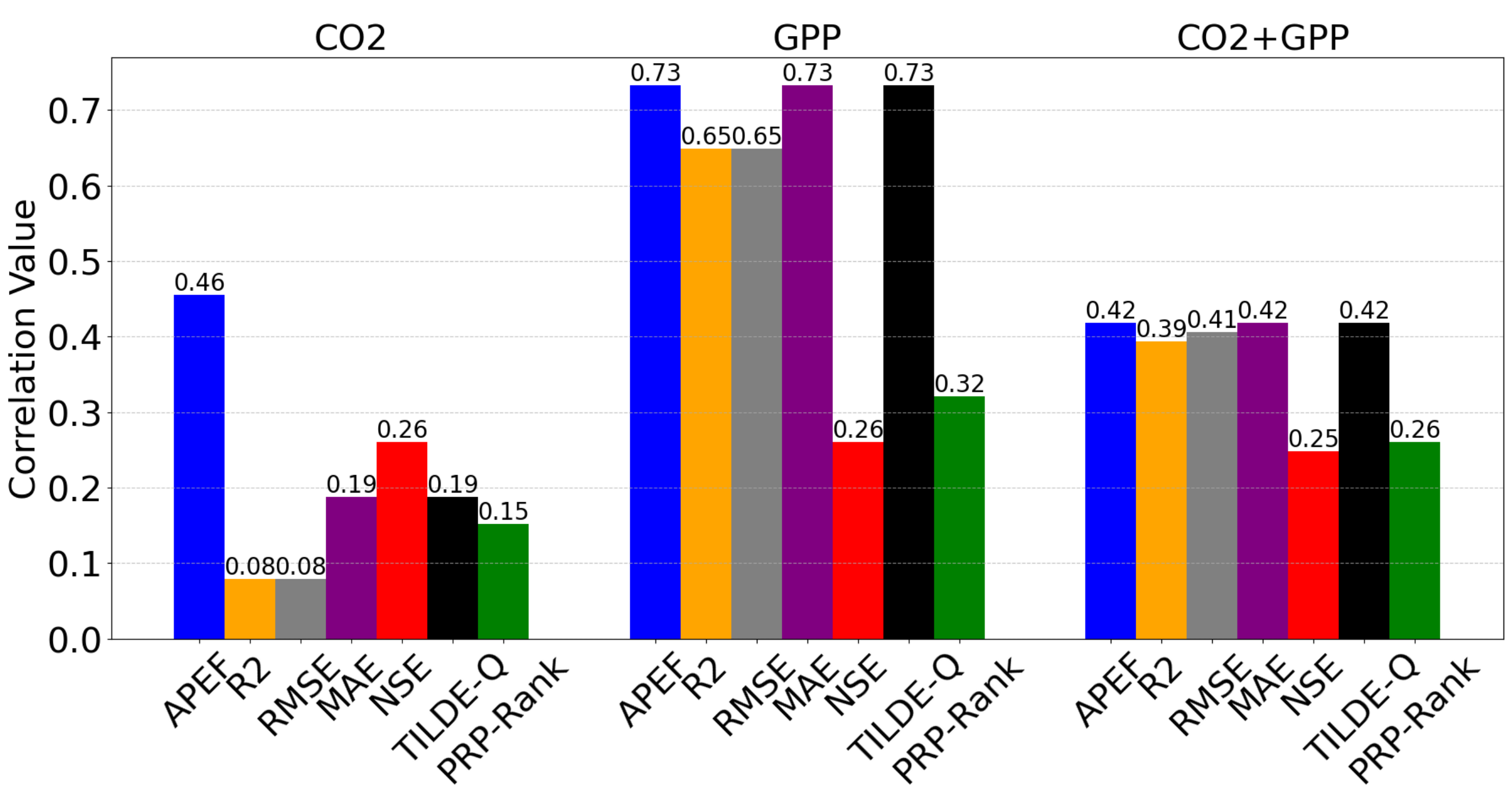}
}
\vspace{-0.2in}
\caption{Correlation performance on the synthetic data using preset weights.}
\label{fig:exp1_bar_pp}
\end{figure}



\begin{figure*} [!h]
\centering
\subfigure[]{ 
\includegraphics[width=0.32\textwidth]{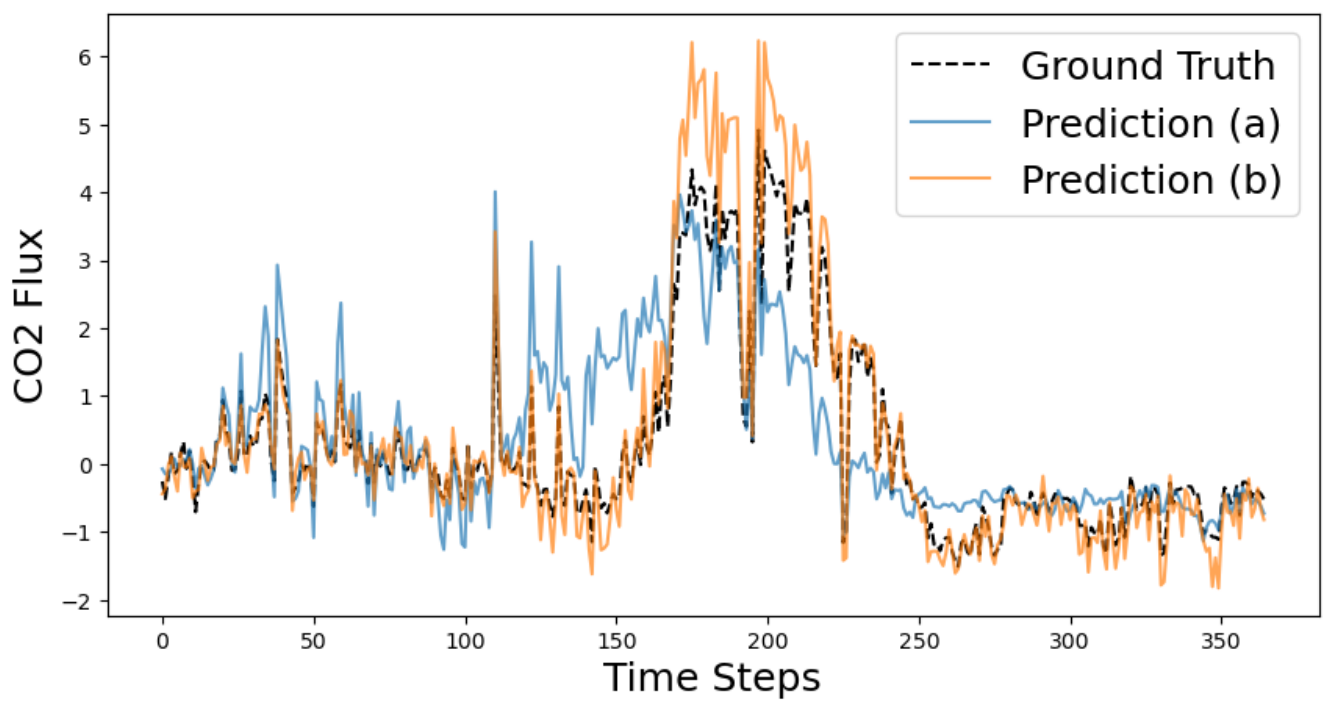}
}
\subfigure[]{ 
\includegraphics[width=0.32\textwidth]{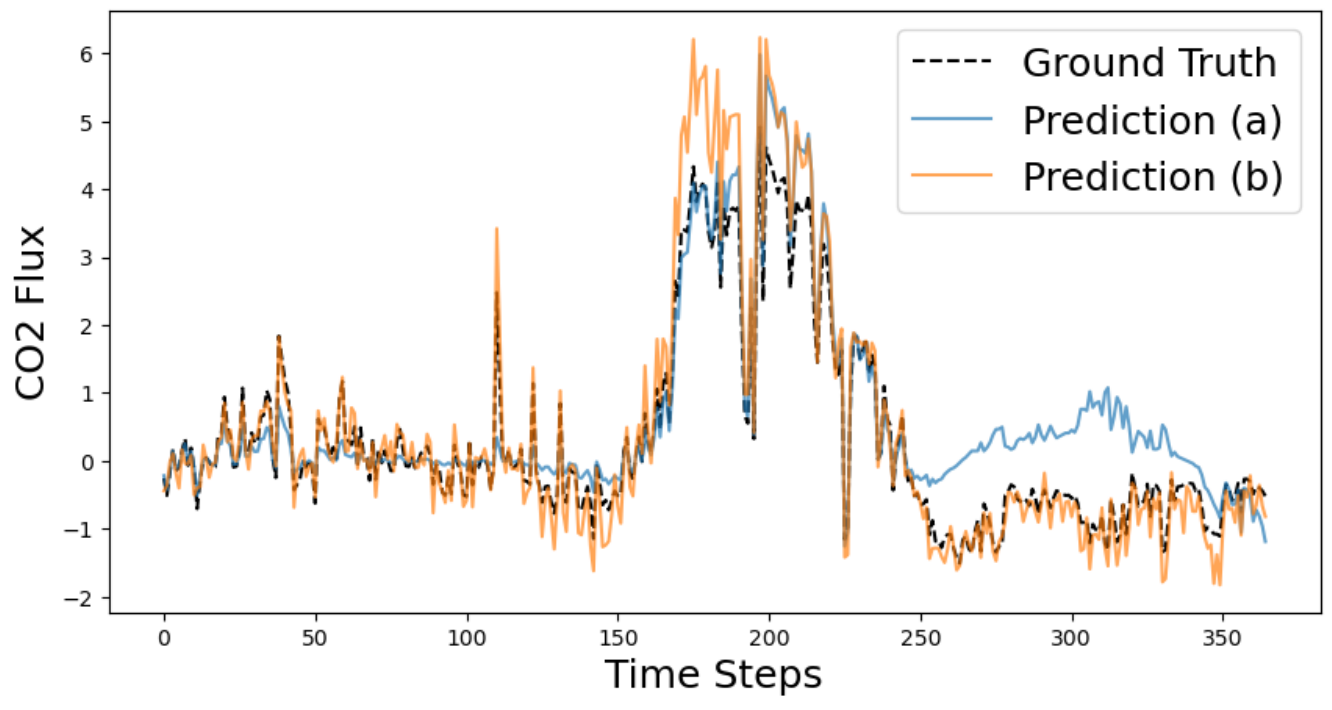}
}
\subfigure[]{ 
\includegraphics[width=0.32\textwidth]{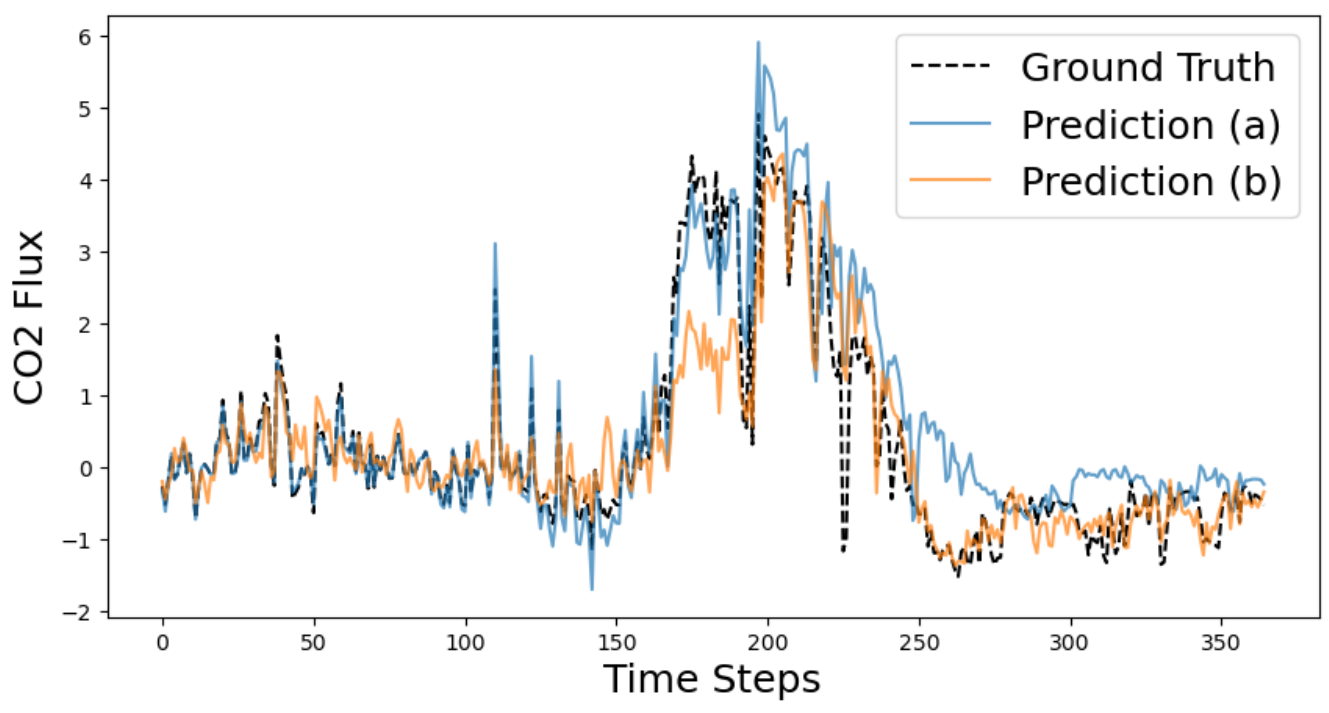}
}
\vspace{-0.2in}
\caption{Examples of ranking mismatch in the Peak period weight setting. In these examples, prediction (a) is better than prediction (b) according to target annotation and APEF, 
but standard metric $R^2$ produces opposite comparison outcome. 
}
\vspace{-0.1in}
\label{fig:mismatch_pp}
\end{figure*}

In this section, we present and analyze the performance of APEF and baseline metrics on the three datasets described in Section \ref{sec:dataset}. The baseline metrics include traditional metrics commonly used for time series regression tasks, including coefficient of determination ($R^2$), root mean square error (RMSE), mean absolute error (MAE), and NS efficiency coefficient (NSE)~\cite{NSE_backup}. We also adopted two evaluation approaches, TILDE-Q \cite{lee2024tildeqtransformationinvariantloss} and PRP-Rank \cite{qin2024largelanguagemodelseffective} as our baselines. TILDE-Q calculates the shape difference between time series, and PRP-Rank also employ LLMs to generate a ranking of predictions by doing pairwise comparison. As the original PRP-Rank is designed to evaluate text generation, we feed the pairwise time series samples and the observation series as text to PRP-Rank to obtain the ranking. 

For the experiment on synthetic dataset and human annotation, we used 10 model predictions as the training set, 5 model predictions as validation set, and 5 model prediction as the testing set. For the experinment on ILAMB dataset, there are 7 model predictions each for training, validation and testing set. We used o3-mini from OpenAI as the LLM for both APEF and for PRP-Rank. All result presented are performances on testing set. More detail about experiment setting can be found in Appendix \ref{sec:appendix-evaluation}.


\subsubsection{Performance on Synthetic Dataset}
Table \ref{tab:exp1} presents the testing performance of APEF and multiple baseline metrics on the synthetic dataset described in Section~\ref{sec:synthetic-dataset}. For this experiment, we use each of the metrics to rank the samples in testing set and calculate the Spearman's correlation between the generated ranking and the ground truth ranking. Here the ground truth ranking is generated based on the base metric score (Section \ref{sec:base-metric}) with the preset weights. 

We can observe from Figure~\ref{fig:exp1_bar_pp} that APEF  outperforms other metrics in the Peak Period weight setting which aims to put more focus on the similarity between prediction and ground truth within the peak period of CO$_2$ flux and GPP. The performance difference  between APEF and some baselines for certain variables  is not that large because the peak period is only about 120 days out of a year.

While other metric treat all period of the time series equally, APEF would extract metrics that focus more on the peak period. For instance, in the resulting policy of CO$_2$ flux, we see a unique metric extracted by LLM  called Peak Period Consistency Score (PPCS), for measuring the length difference of the peak period. This is defined by the LLM  as follow:
\begin{equation}
\small
    \begin{aligned}
        PPCS = 1 - \frac{|\text{Length Peak Period}_\text{prd} - \text{Length Peak Period}_\text{tgt}| }{\text{Length Peak Period}_\text{tgt}}, 
    \end{aligned}
\end{equation}
where $\text{Length Peak Period}_\text{tgt}$ and $\text{Length Peak Period}_\text{prd}$ represent the length of peak period detected from the reference series $Y^i$ and the predicted series $P^i$.  

We can also observe that the performance of APEF outperforms or matches the performance of other metrics in both the Slope \& Curvature and  Amplitude settings. This is also expected because the traditional metrics are designed to measure the overall errors between prediction and ground truth time series. For these two settings, we observed that APEF included traditional metrics as rules in the resulting policy. For example, we see the following two scoring rules in the final policy of the CO$_2$ flux in Amplitude setting: $- \text{RMSE}: 2 \,\,\text{points}; - \text{MAE}: 2 \,\,\text{points}$. 

\subsubsection{Performance on Human Expert Alignment}
In this experiment, we evaluate the alignment between the predicted ranking and the target ranking derived from human annotators' pairwise comparisons of model predictions. 
We show the correlation between expert ranking and metric rankings in Table \ref{tab:exp2} (left). 
We can observe that APEF outperforms other metrics for this settings. The resulting policies in this experiment 
include novel rules extracted by the LLM. One example metric is about the proportion of time steps with large errors, and is defined as $N_\text{violate} / T$, where  $N_\text{violate} = \text{Count}\{t \in \{1, ..., T\} \,\,\text{s.t.} \,\, |Y^i_t - P^i_t| > 1.0\}$. 


\subsubsection{Performance on the ILAMB dataset}
This test uses the ILAMB scoring system, described in Section \ref{sec:ilamb-dataset}, to rank a set of seven CIMP6 models, and uses this ranking as the target ranking. The ILAMB dataset does not provide GPP and CO$_2$ data for the same instances so we only evaluate them separately. Since the scoring function used in ILAMB is a simple combination of traditional metrics like 
bias score and RMSE score, most traditional metrics tend  to produce high correlation performance (>0.85). 
Table \ref{tab:exp2} (right) only shows the performance of APEF and the baselines TILDE-Q and PRP-Rank. TILDE-Q is essentially similar to the overall error metric (e.g., RMSE) and thus gets high correlation.  It is worth noticing that the proposed framework still yield comparable result even when the target score does not explicitly consider special temporal characteristics in time series. 
More importantly, we observe that the final policies learned by APEF consists entirely of 
traditional metrics. The metric section of the policy includes ``Mean Absolute Error (MAE), Root Mean Square Error (RMSE), Variance Explained (R$^2$), Pearson Correlation Coefficient". 

\subsection{Case Study}
Figure \ref{fig:mismatch_pp} shows  examples where the target score based on the base metric in the Peak Period setting ranks model (a) to be higher than model (b). However,  the traditional metric, such as $R^2$,  suggests that   model (b) is better. We  observe that the  predictions  (a) is more closely aligned to the ground truth time series inside peak period (Day 160 to 240) while the predictions  (b) is better aligned to ground truth series outside of peak period. This is a common scenario in CO$_2$ flux prediction where models can predict reasonably well in normal days but when there  is a period of abrupt changes, the models perform poorly. For this example pair, APEF successfully identifies model (a) to be the better model, because it extracts metrics that focus on the peak period. We observe the following metric in the resulting policy: ``Peak Period Deviation (PPD): Quantifies how close the observed series peak period  is to the reference  via an exponential decay." Even though this is very different from how the target score defines the similarity within peak period, but APEF successfully learns to focus on the peak period. 

\section{Conclusion}
This paper proposes a new framework APEF for evaluating time series output of ecological models. Our results on both synthetic and real datasets highlight several key findings: (1) APEF effectively captures complex assessment criteria (e.g., those provided by human expert annotators), as evidenced by its high correlation with target scores.  (2) APEF provides interpretable policies, with extracted metrics reflecting key assessment priorities. (3) APEF adapts well to different evaluation settings, such as the synthetic data with varying weight priorities within the base metric. 
APEF builds foundation for  automated and interpretable evaluation for ecological time series. We also anticipate it to serve as cornerstone for broader assessment of complex data across various scientific domains. 
Future work includes extending APEF to accommodate a larger set of interdependent variables and incorporate user-specified assessment priorities in the generated policies.

\bibliography{reference, shengyu, Xiaowei, Qi}
\bibliographystyle{ACM-Reference-Format}

\appendix

\section{Policy examples}
\subsection{Policy example for two variables} \label{sec:appendix-2f}
 
\begin{figure}[h]
\centering
\includegraphics[width=0.45\textwidth]{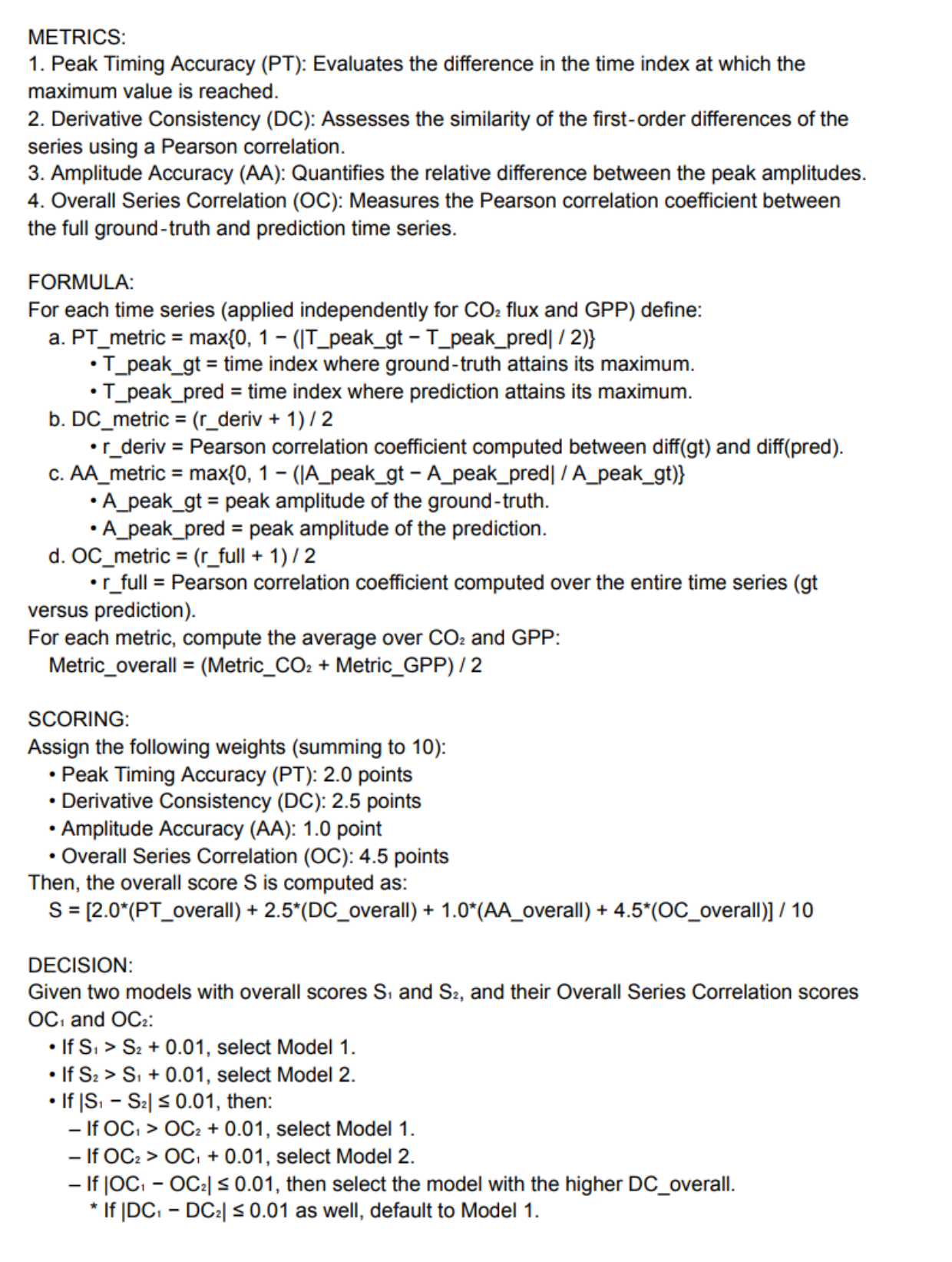}
\caption{ Example policy that reflects the correlation between two variables.}
\label{fig:exp1_policy_2f}
\end{figure}

 Fig.~\ref{fig:exp1_policy_2f} shows an example of a policy learned for evaluating a pair of time series for two different variables.

\section{Prompt Example}
\subsection{Weight Optimization} \label{sec:appenix:WO}
\begin{figure}[h]
\centering
\includegraphics[width=0.45\textwidth]{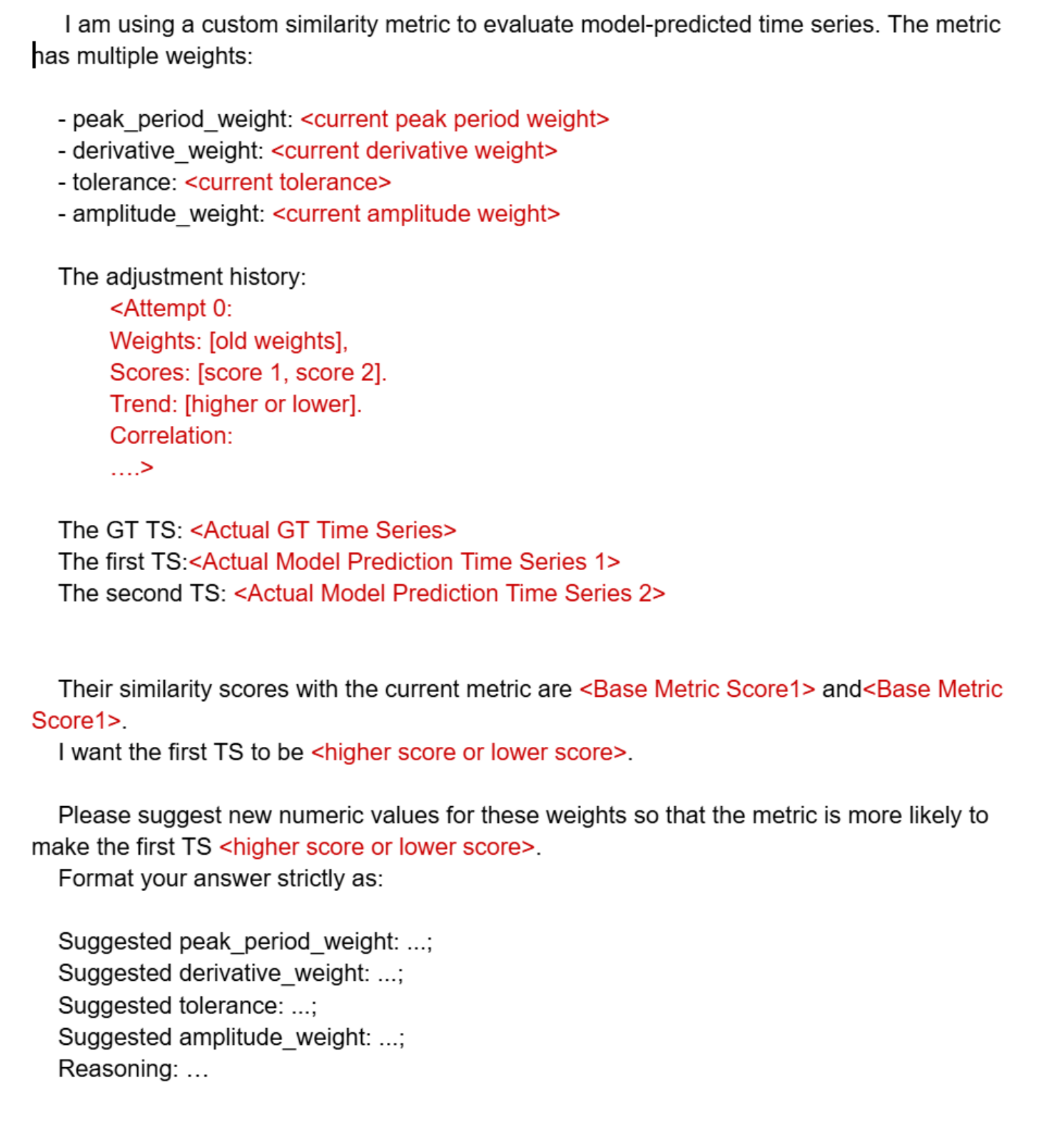}
\caption{ Example prompt for doing a single iteration in Weight Optimization. Color red represent actual variables we need to input.}
\label{fig:exp_prompt_WO}
\end{figure}

Example prompt that we used to do weight optimization is shown in Figure \ref{fig:exp_prompt_WO}.

\section{Dataset}

\subsection{Human Annotation} \label{sec:anno}
An example annotation task we gave to the annotators is shown in Figure \ref{fig:exp_annotation_appendix}.
\begin{figure}[h]
\centering
\includegraphics[width=0.45\textwidth]{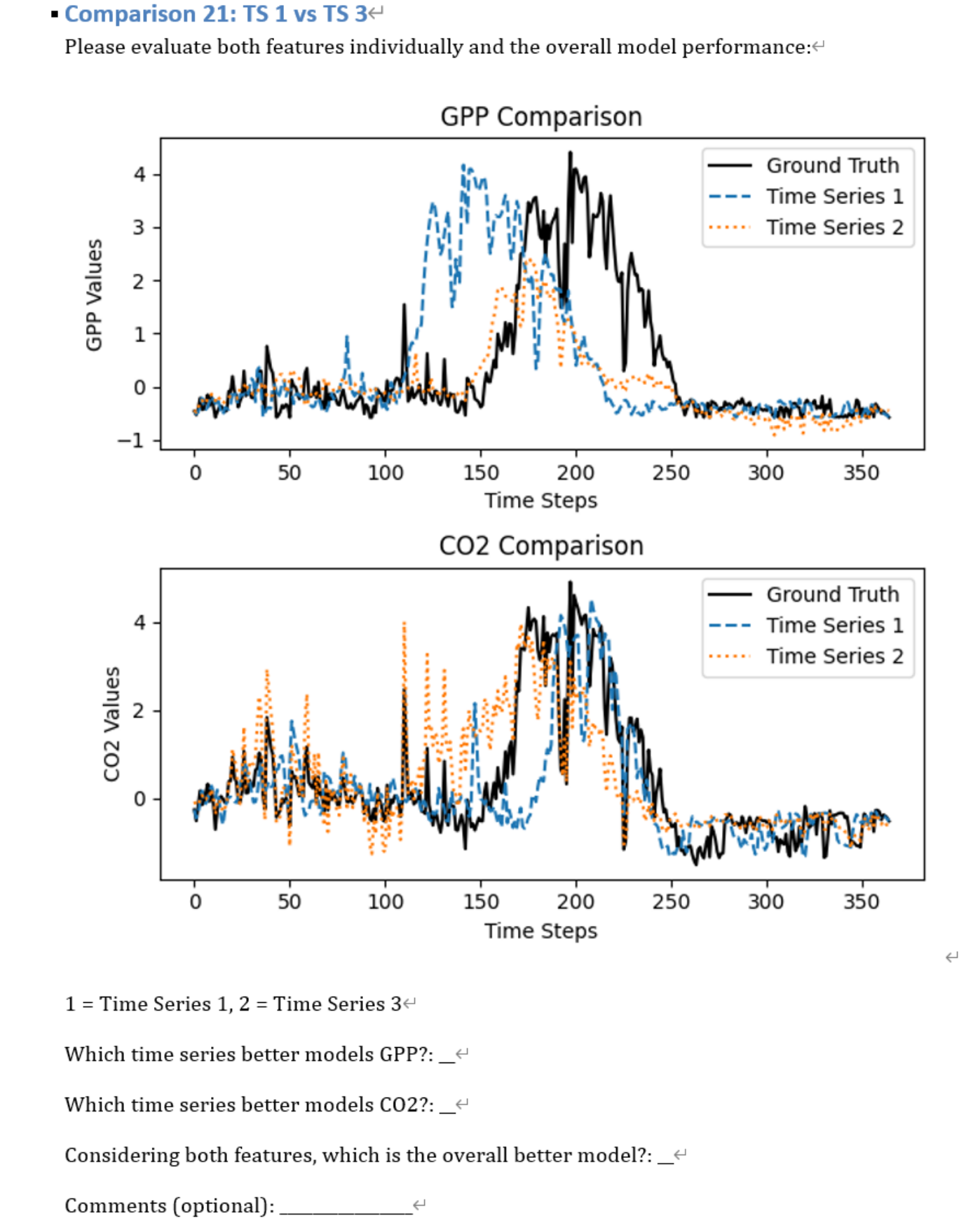}
\caption{ Example annotation task.}
\label{fig:exp_annotation_appendix}
\end{figure}

\subsection{ILAMB Dataset}\label{sec:appendix-ilamb}
Since our evaluation focused on single-site, single-year predictions, we excluded the Spatial Distribution Score and Interannual Variability Score as they require multi-site or multi-year data.The Bias Score measures the average difference between model predictions and observations, normalized by the observational mean. The RMSE Score evaluates the root mean square error, normalized by the observational standard deviation. The Seasonal Cycle Score assesses how well the model captures the timing and magnitude of seasonal patterns by comparing the mean seasonal cycles. Each component score is transformed to a [0,1] range. 

The CMIP6 models we used include CanESM5, ACCESS-ESM1-5, IPSL-CM6A-LR, UKESM1-0-LL, MPI-ESM1-2-HR, CESM2, and NorESM2-LM \cite{gmd-12-4823-2019, ACCESS_ESM, IPSL, UKESM1, gmd-12-3241-2019, CESM, gmd-13-6165-2020}. 

\section{Evaluation}\label{sec:appendix-evaluation}
For all the experiments, we first calculate all spearman correlation between target ranking of the model predictions in the testing set and the scores from metrics in the baseline, including RMSE, $R^2$, NSE, MAE, and TILDE-Q. Than we generate the ranking of the model predictions using PRP-Rank which we set K=2. For training APEF with the model predictions in the training set, we first randomly initiate a set of weight which we would plug in to the base metric. Second, we do 10 iterations of pairwise comparison described in the weight optimization process as warm up. The reason we do warm up iterations is that weight initiated randomly can often yield correlation close to 0 with the target ranking, which provide limited information to LLMs in the policy extraction process. After warming up, we do another 10 iterations of weight optimization. For each iteration after the warm up, we also perform policy extraction process. The generated policy then use validation set to check the correlation with target ranking in the validation set. If the correlation on validation set is improved comparing to the policy extracted in previous iteration, then we accept this policy as the new policy. It is worth mentioning that we use empty policy as a start and since APEF iteratively improve the policy, the quality of the first extracted policy is critical. As a result, the first exctracted policy is only accepted when it reached a positive correlation on validation set. 

For PRP-Rank in the baseline and all the steps in APEF that involve using LLM, we used o3-mini from OpenAI as the LLM. In the code, we also provide the option to use LLama 3.2.

\begin{figure}[h]
\centering
\includegraphics[width=0.45\textwidth]{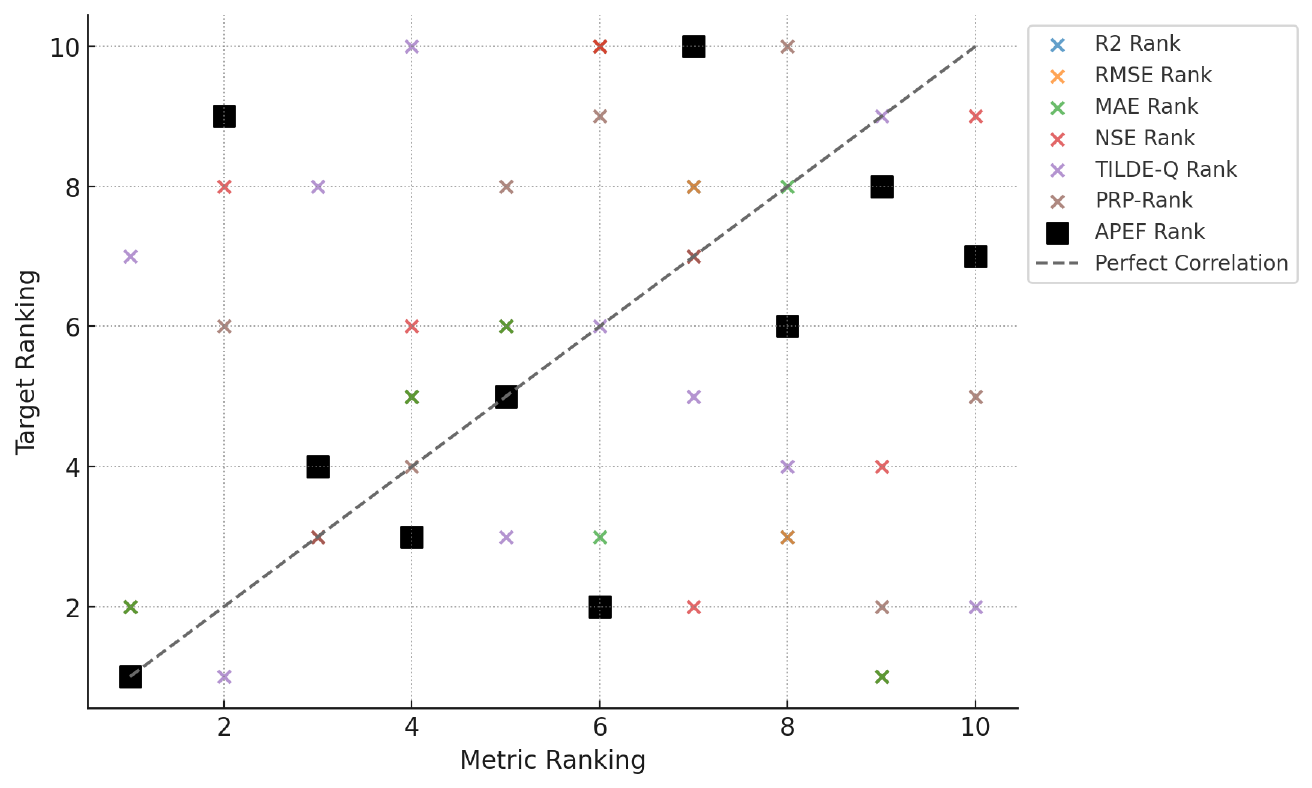}
\caption{ Detailed model ranking using validation and testing set of synthetic data as a scatter plot. Y values are target ranking from base metric using peak period weight setting. X values are rankings from different metric. Perfect correlation is the diagonal line in color grey.}
\label{fig:exp_annotation_appendix}
\end{figure}


\begin{table}[!t]
\small
\caption{Correlation with expert annotations (left) and the ILAMB scores (right) on winter focus. }
\begin{tabular}{c|clclcl|cc}
\hline
 & \multicolumn{6}{c|}{Annotation} & \multicolumn{2}{c}{ILAMB} \\ \hline
      & \multicolumn{2}{c}{CO$_2$}   & \multicolumn{2}{c}{GPP} & \multicolumn{2}{c|}{GPP + CO$_2$} & {CO$_2$} & GPP\\ \hline
$R^2$ & \multicolumn{2}{c}{0.373} & \multicolumn{2}{c}{0.443}  & \multicolumn{2}{c|}{0.017
} & - & -            \\
RMSE& \multicolumn{2}{c}{0.373} & \multicolumn{2}{c}{0.443}  & \multicolumn{2}{c|}{0.117
}   & - & -          \\
MAE& \multicolumn{2}{c}{0.357} & \multicolumn{2}{c}{0.429}  & \multicolumn{2}{c|}{0.100
}          & - & -   \\
NSE& \multicolumn{2}{c}{0.258} & \multicolumn{2}{c}{0.370}  & \multicolumn{2}{c|}{0.283
}     & - & -        \\
TILDE-Q & \multicolumn{2}{c}{0.357} & \multicolumn{2}{c}{0.429}  & \multicolumn{2}{c|}{0.100
} & \textbf{-} & \textbf{-}            \\
PRP-Rank& \multicolumn{2}{c}{0.598} & \multicolumn{2}{c}{0.625}  & \multicolumn{2}{c|}{0.383
} & - & -            \\ \hline
APEF before & \multicolumn{2}{c}{0.761} & \multicolumn{2}{c}{0.616}    & \multicolumn{2}{c|}{0.417}  & - & - \\ \hline    
APEF after & \multicolumn{2}{c}{0.782} & \multicolumn{2}{c}{0.633}    & \multicolumn{2}{c|}{0.567}  & - & - \\ \hline      
\end{tabular}
\label{tab:exp2}
\end{table}

%


\begin{table*}[!t]
\small
\caption{Correlation with simulated weight sets. Each column (Peak Period, Slope \& Curvature, and Amplitude) represent a preset scenario where the indicated aspect is more valued than others. Gradient Version}
\begin{tabular}{c|clclc|ccc|ccc}
\hline
\multicolumn{1}{l|}{} & \multicolumn{5}{c|}{Peak Period}                                                          & \multicolumn{3}{c|}{Slope \& Curvature} & \multicolumn{3}{c}{Amplitutde} \\ \hline
                      & \multicolumn{2}{c}{CO$_2$}         & \multicolumn{2}{c}{GPP}            & GPP + CO$_2$    & CO$_2$     & GPP     & GPP + CO$_2$     & CO$_2$  & GPP  & GPP + CO$_2$  \\ \hline
$R^2$                 & \multicolumn{2}{c}{0.067}          & \multicolumn{2}{c}{0.139
}          & -
&            -&         -
&                  -
&         -
&      -
&               -
\\
RMSE                  & \multicolumn{2}{c}{0.067}          & \multicolumn{2}{c}{0.139
}          & -
&            
&         -&                  
&         
&      
&     
\\
MAE                   & \multicolumn{2}{c}{0.067}          & \multicolumn{2}{c}{0.127
}          & -
&            -&         -
&                  -
&         -
&      -
&      -
\\
NSE                   & \multicolumn{2}{c}{0.407}          & \multicolumn{2}{c}{0.297
}          & -
&            
&         -&                  -
&         -
&      -
&               -
\\
TILDE-Q               & \multicolumn{2}{c}{0.067}          & \multicolumn{2}{c}{0.127
}          & -
&            -&         -
&                  
&         
&      
&               
\\
PRP-Rank              & \multicolumn{2}{c}{0.321}          & \multicolumn{2}{c}{0.067
}            & -
&            
&         -&                  -
&         -
&      -
&               -
\\ \hline
APEF                  & \multicolumn{2}{c}{\textbf{0.452}} & \multicolumn{2}{c}{0.300} & -
&            -&         -
&    -
&         -
&      -
&              
-
\\ \hline 
APEF (Gradient)                 & \multicolumn{2}{c}{0.439} & \multicolumn{2}{c}{\textbf{0.337}} & -&            
&         -&    
&         
&      
&              

\\ \hline 

\end{tabular}
\label{tab:exp1}
\vspace{-.15in}
\end{table*}

\end{document}